\documentclass[11pt]{article}

\usepackage[preprint]{acl}

\usepackage{times}
\usepackage{latexsym}

\usepackage[T1]{fontenc}

\usepackage[utf8]{inputenc}

\usepackage{microtype}

\usepackage{inconsolata}

\usepackage{graphicx}
\usepackage{amsmath}
\usepackage{array}
\usepackage{booktabs}
\usepackage{algorithm}
\usepackage{algorithmic}
\usepackage{hyperref}
\usepackage{amssymb}
\usepackage{bm}
\usepackage{multirow}

\title{NeuroLoRA: Context-Aware Neuromodulation for Parameter-Efficient Multi-Task Adaptation}

\author{
    \textbf{Yuxin Yang\textsuperscript{1}} \quad
    \textbf{Haoran Zhang\textsuperscript{1}} \quad
    \textbf{Mingxuan Li\textsuperscript{2}} \quad
    \textbf{Jiachen Xu\textsuperscript{1}} \quad
    \textbf{Ruoxi Shen\textsuperscript{2}} \\
    \textbf{Zhenyu Wang\textsuperscript{1}} \quad
    \textbf{Tianhao Liu\textsuperscript{2}} \quad
    \textbf{Siqi Chen\textsuperscript{1}} \quad
    \textbf{Weilin Huang\textsuperscript{2}}
    \\
    \\
    \textsuperscript{1}Shanghai University \quad
    \textsuperscript{2}Fudan University
}

\begin{document}
\maketitle
\begin{abstract}
Parameter-Efficient Fine-Tuning (PEFT) techniques, particularly Low-Rank Adaptation (LoRA), have become essential for adapting Large Language Models (LLMs) to downstream tasks. While the recent FlyLoRA framework successfully leverages bio-inspired sparse random projections to mitigate parameter interference, it relies on a static, magnitude-based routing mechanism that is agnostic to input context. In this paper, we propose \textbf{NeuroLoRA}, a novel Mixture-of-Experts (MoE) based LoRA framework inspired by biological neuromodulation---the dynamic regulation of neuronal excitability based on context. NeuroLoRA retains the computational efficiency of frozen random projections while introducing a lightweight, learnable neuromodulation gate that contextually rescales the projection space prior to expert selection. We further propose a Contrastive Orthogonality Loss to explicitly enforce separation between expert subspaces, enhancing both task decoupling and continual learning capacity. Extensive experiments on MMLU, GSM8K, and ScienceQA demonstrate that NeuroLoRA consistently outperforms FlyLoRA and other strong baselines across single-task adaptation, multi-task model merging, and sequential continual learning scenarios, while maintaining comparable parameter efficiency.
\end{abstract}

\section{Introduction}

The paradigm of pre-training followed by task-specific fine-tuning has become the dominant approach for deploying Large Language Models (LLMs) \cite{brown2020language, achiam2023gpt, grattafiori2024llama}. However, Full Fine-Tuning (FFT) of models with billions of parameters incurs prohibitive computational and storage costs, motivating the development of Parameter-Efficient Fine-Tuning (PEFT) methods \cite{houlsby2019parameter, lester2021power, li2021prefix}. Among these, Low-Rank Adaptation (LoRA) \cite{hu2022lora} has emerged as the de facto standard, freezing the pre-trained weights and injecting trainable low-rank decomposition matrices into the model's linear layers.

Despite its efficiency, standard LoRA assumes a single global low-rank update shared across all inputs. This design leads to \textit{intra-task interference}---where diverse input patterns within a single task compete for the same limited rank capacity---and \textit{inter-task interference} when merging adapters trained on different tasks \cite{yu2020gradient, yadav2023ties, zou2025flylora}. To address these limitations, Mixture-of-Experts (MoE) architectures have been integrated into the LoRA framework \cite{dou2024loramoe, wu2024mixture, li2024mixlora}, enabling conditional activation of task-specific parameter subsets.

A recent advancement, \textbf{FlyLoRA} \cite{zou2025flylora}, draws inspiration from the fruit fly olfactory circuit \cite{dasgupta2017neural} to replace trainable routers with a frozen sparse random projection matrix. By selecting experts based on the magnitude of the projected input (a ``Winner-Take-All'' mechanism analogous to Kenyon Cell activation), FlyLoRA achieves strong task decoupling and enables training-free model merging.

However, FlyLoRA's reliance on a fixed random projection represents a purely \textit{structural} imitation of biological neural circuits, overlooking a critical functional component: \textbf{neuromodulation} \cite{marder2012neuromodulation}. In biological brains, neural responses are dynamically modulated by neurotransmitters (e.g., dopamine, acetylcholine) that alter neuronal sensitivity based on internal states and external context \cite{doya2002metalearning}. FlyLoRA's routing is reactive to the current token's magnitude profile but agnostic to the broader semantic context. This static routing can be suboptimal for tasks requiring contextual reasoning, where the appropriate expert specialization depends on the meaning of preceding tokens rather than the current token's magnitude alone.

To bridge this gap, we propose \textbf{NeuroLoRA} (Neuromodulated Low-Rank Adaptation). NeuroLoRA preserves the computational efficiency and merging compatibility of FlyLoRA's frozen sparse projection while augmenting it with a \textit{Context-Aware Neuromodulation Gate}. This lightweight gating component dynamically rescales the axes of the random projection space based on input context, effectively modulating the routing decision boundary without altering the projection matrix itself. Furthermore, we introduce a \textit{Contrastive Orthogonality Loss} ($\mathcal{L}_{orth}$) that explicitly enforces separation between expert subspaces, providing stronger guarantees than the probabilistic near-orthogonality of random projections alone. This design additionally confers natural advantages for continual learning, as the enforced subspace separation mitigates catastrophic forgetting \cite{kirkpatrick2017overcoming} when the model is sequentially adapted to new tasks.

Our contributions are summarized as follows:
\begin{itemize}
    \item We identify the limitation of static magnitude-based routing in FlyLoRA and propose a biologically motivated solution grounded in the principle of neuromodulation.
    \item We introduce NeuroLoRA, which incorporates a lightweight context-aware gating mechanism to dynamically modulate the frozen projection space, enabling context-sensitive expert activation while preserving training-free merging compatibility.
    \item We propose a Contrastive Orthogonality Loss that explicitly penalizes expert subspace overlap, enhancing task decoupling for both multi-task merging and continual learning.
    \item Extensive experiments on three benchmarks (MMLU, ScienceQA, GSM8K) demonstrate that NeuroLoRA outperforms FlyLoRA and other strong baselines in single-task, multi-task merging, and continual learning settings.
\end{itemize}

\section{Related Work}

\subsection{Parameter-Efficient Fine-Tuning}

PEFT aims to adapt pre-trained LLMs to downstream tasks while updating only a small fraction of parameters. Early approaches include Adapter layers \cite{houlsby2019parameter}, which insert bottleneck modules between Transformer \cite{vaswani2017attention} layers, and Prefix-Tuning \cite{li2021prefix}, which prepends learnable continuous tokens to the input. Prompt Tuning \cite{lester2021power} further simplifies this by learning soft prompts in the input embedding space.

\citet{hu2022lora} introduced LoRA, which injects low-rank matrices $B \in \mathbb{R}^{d_{out} \times r}$ and $A \in \mathbb{R}^{r \times d_{in}}$ into frozen linear layers, where $r \ll \min(d_{in}, d_{out})$. Subsequent work has focused on improving rank allocation and training dynamics. \textbf{AdaLoRA} \cite{zhang2023adalora} dynamically prunes singular values via importance scoring to allocate rank budget across layers. \textbf{QLoRA} \cite{dettmers2024qlora} combines 4-bit quantization of the base model with LoRA adapters, substantially reducing memory footprint. \textbf{DoRA} \cite{liu2024dora} decomposes the weight update into magnitude and directional components to improve learning stability. While effective for single-task adaptation, these methods are susceptible to parameter interference in multi-task and continual learning settings \cite{zou2025flylora, wang2023orthogonal}.

\subsection{Mixture-of-Experts for LoRA}

Mixture-of-Experts (MoE) architectures conditionally activate subsets of parameters based on input characteristics \cite{shazeer2017outrageously, fedus2022switch, lepikhin2021gshard}, enabling increased model capacity without proportional computational cost. Integrating MoE with LoRA has emerged as a principled approach to multi-task adaptation. \textbf{LoRAMoE} \cite{dou2024loramoe} employs a localized balancing constraint to prevent expert collapse during instruction tuning. \textbf{MoLE} \cite{wu2024mixture} and \textbf{MixLoRA} \cite{li2024mixlora} utilize top-$k$ gating networks to route tokens to different LoRA expert modules. \citet{zadouri2023pushing} investigated the parameter efficiency limits of MoE-LoRA configurations. \textbf{AdaMix} \cite{wang2022adamix} randomly mixes adapter experts during training for improved generalization.

However, these methods rely on explicit, trainable router networks (typically linear layers followed by softmax), which introduce additional learnable parameters and---more critically---make training-free model merging infeasible since the routing logic is learned and task-specific. NeuroLoRA differs fundamentally by retaining a frozen projection for routing while introducing dynamic modulation.

\subsection{Bio-Inspired Representations and Model Merging}

The fruit fly olfactory circuit has inspired efficient locality-sensitive hashing algorithms \cite{dasgupta2017neural}. The core principle---projecting inputs into a high-dimensional sparse space to separate overlapping patterns---was applied to LoRA by \textbf{FlyLoRA} \cite{zou2025flylora}, which treats the down-projection matrix $A$ as a frozen sparse random projection and uses magnitude-based Winner-Take-All activation for expert selection. This design inherently supports \textit{Model Merging} techniques such as Task Arithmetic \cite{ilharco2023editing} and TIES-Merging \cite{yadav2023ties}, due to the approximate orthogonality of high-dimensional random vectors.

Beyond task decoupling and merging, recent work has revealed that the structural properties of the fly olfactory circuit---in particular, its sparse, high-dimensional, and approximately orthogonal random projections---naturally mitigate the stability-plasticity dilemma central to continual learning \cite{zou2025structural}. \textbf{Fly-CL} \cite{zou2026flycl} further extended this insight into a practical continual representation learning framework, demonstrating that fly-inspired decorrelation mechanisms can significantly reduce inter-task interference while maintaining training efficiency.

Our work identifies a critical gap in FlyLoRA: while it imitates the \textit{structure} (fixed connectivity) of the fly olfactory circuit, it omits the \textit{functional dynamics} of biological neural systems. By introducing a neuromodulation gate inspired by the role of modulatory neurotransmitters in biological brains \cite{marder2012neuromodulation, doya2002metalearning}, NeuroLoRA combines the structural stability of random projections with contextual adaptability.

\subsection{Continual Learning with PEFT}

Continual learning (CL) seeks to sequentially learn new tasks without catastrophically forgetting previously acquired knowledge \cite{kirkpatrick2017overcoming, li2017learning, mccloskey1989catastrophic}. Classical CL strategies include regularization-based methods such as Elastic Weight Consolidation (EWC) \cite{kirkpatrick2017overcoming} and Learning without Forgetting (LwF) \cite{li2017learning}, replay-based methods \cite{rebuffi2017icarl, chaudhry2019episodic}, and architecture-based methods that allocate dedicated parameters per task \cite{rusu2016progressive}.

The intersection of PEFT and continual learning has attracted increasing attention. \citet{razdaibiedina2023progressive} proposed Progressive Prompts, which sequentially append new soft prompts while freezing previous ones. \textbf{O-LoRA} \cite{wang2023orthogonal} constrains each new task's LoRA subspace to be orthogonal to previously learned subspaces, directly addressing inter-task interference. \citet{biderman2024lora} empirically demonstrated that LoRA provides a degree of implicit regularization against forgetting due to its low-rank constraint. InfLoRA \cite{liang2024inflora} further forms LoRA parameters within the subspace spanned by prior task representations.

From a bio-inspired perspective, \citet{zou2025structural} provided theoretical and empirical evidence that the sparse, expansive coding of the fly olfactory circuit inherently alleviates catastrophic forgetting by projecting task representations into nearly orthogonal subspaces. \citet{zou2026flycl} built upon this insight to develop Fly-CL, a fly-inspired continual learning framework that achieves efficient decorrelation across sequential tasks with reduced training overhead. NeuroLoRA's design extends this line of work: the frozen projection matrix $A$ provides the same structural stability, the context-aware modulation gate further adapts expert selection to task-specific patterns, and the Contrastive Orthogonality Loss explicitly enforces non-overlapping expert subspaces, reducing destructive interference between sequentially learned tasks.

\section{Methodology}

\subsection{Preliminaries: Revisiting FlyLoRA}

Standard LoRA \cite{hu2022lora} approximates the weight update $\Delta W \in \mathbb{R}^{d_{out} \times d_{in}}$ for a frozen pre-trained weight matrix $W_0$ with two low-rank matrices: $B \in \mathbb{R}^{d_{out} \times r}$ and $A \in \mathbb{R}^{r \times d_{in}}$, where rank $r \ll \min(d_{in}, d_{out})$. The modified forward pass is:
\begin{equation}
    y = W_0 x + \frac{\alpha}{r} B A x
\end{equation}
where $\alpha$ is a scaling hyperparameter.

FlyLoRA \cite{zou2025flylora} reinterprets this decomposition as a Rank-Wise Mixture of Experts, drawing on the sparse, expansive coding principle of the fruit fly olfactory system \cite{dasgupta2017neural}. It introduces two key modifications:
\begin{enumerate}
    \item \textbf{Frozen Sparse Projection.} The down-projection matrix $A$ is initialized as a sparse random matrix with sparsity ratio $\rho$ and remains frozen throughout training. Each element $A_{ij}$ is drawn from $\{0, +1, -1\}$ with probabilities $\{1-\rho, \rho/2, \rho/2\}$. This mimics the fixed, random connections from Projection Neurons (PNs) to Kenyon Cells (KCs) in the fly olfactory circuit.
    \item \textbf{Implicit Magnitude-Based Routing.} For an input $x \in \mathbb{R}^{d_{in}}$, FlyLoRA computes the projection $h = Ax \in \mathbb{R}^r$. The top-$k$ indices in the absolute values $|h|$ determine the set of activated experts $\mathcal{I}_{active}$, mimicking the Winner-Take-All mechanism observed in KCs.
\end{enumerate}

The output is then computed as:
\begin{equation}
    \Delta y = \sum_{i \in \mathcal{I}_{active}} B_{:,i} \cdot h_i
\end{equation}

The critical limitation of this design lies in its \textit{context-agnostic} routing. The expert selection for a token $x$ depends solely on its intrinsic vector representation and the fixed projection $A$. For instance, the token ``bank'' should ideally activate different experts in the contexts of ``river bank'' versus ``investment bank,'' but a static, magnitude-based router cannot distinguish between these cases since the token embeddings at the LoRA injection point carry limited contextual differentiation.

\subsection{NeuroLoRA: Dynamic Modulation of Static Projections}

We draw inspiration from \textbf{neuromodulation} in biological neural systems \cite{marder2012neuromodulation}. Neuromodulators such as dopamine and acetylcholine do not directly transmit primary sensory information; instead, they alter the excitability and gain of target neurons, effectively changing how those neurons respond to the same input depending on the organism's internal state or external context \cite{doya2002metalearning}.

NeuroLoRA simulates this mechanism by introducing a lightweight, context-aware gate that dynamically modulates the static projection space of matrix $A$.

\paragraph{Context-Aware Neuromodulation Gate.}
For each input token $x \in \mathbb{R}^{d_{in}}$, we generate a modulation vector $m_x \in \mathbb{R}^r$ via a bottleneck gating network $E_\phi$:
\begin{equation}
    m_x = \sigma\!\left(W_2 \cdot \text{GELU}(W_1 x)\right) \odot \gamma + \beta
    \label{eq:gate}
\end{equation}
where $W_1 \in \mathbb{R}^{d_h \times d_{in}}$ and $W_2 \in \mathbb{R}^{r \times d_h}$ are learnable projections with bottleneck dimension $d_h \ll d_{in}$, $\sigma(\cdot)$ denotes the sigmoid function, and $\odot$ is the Hadamard product. The learnable parameters $\gamma \in \mathbb{R}^r$ and $\beta \in \mathbb{R}^r$ are initialized to $\mathbf{1}$ and $\mathbf{0}$ respectively, ensuring that $m_x \approx \mathbf{1}$ at initialization. This guarantees that NeuroLoRA reduces to FlyLoRA at the start of training, providing a stable initialization.

\paragraph{Modulated Expert Selection.}
The expert selection is performed on the dynamically re-weighted projection. Instead of computing $h = Ax$ directly, we apply the modulation:
\begin{equation}
    h' = (Ax) \odot m_x
    \label{eq:modulated}
\end{equation}
This operation can be understood as dynamically rescaling the axes of the $r$-dimensional random projection space. Dimensions deemed relevant by the context (via $m_x$) are amplified, increasing their probability of being selected by the top-$k$ operation, while contextually irrelevant dimensions are suppressed.

The active expert set is then determined by:
\begin{equation}
    \mathcal{I}_{active} = \text{TopK}\!\left(|h'|, k\right)
\end{equation}
and the final low-rank update is:
\begin{equation}
    \Delta y = \sum_{i \in \mathcal{I}_{active}} B_{:,i} \cdot h'_i
    \label{eq:output}
\end{equation}

Crucially, the projection matrix $A$ remains frozen and sparse throughout training. Only the gating parameters $\{W_1, W_2, \gamma, \beta\}$ and the up-projection matrix $B$ are updated, preserving both parameter efficiency and merging compatibility.

\subsection{Contrastive Orthogonality Loss}

FlyLoRA relies on the Johnson--Lindenstrauss lemma \cite{johnson1984extensions}, which guarantees that random projections approximately preserve pairwise distances. In high dimensions, random vectors are nearly orthogonal with high probability. However, this is a probabilistic guarantee that weakens for practical, finite ranks (e.g., $r = 32$), where non-negligible correlations between expert columns of $B$ can arise during training.

To strengthen subspace separation from probabilistic to explicitly enforced, we introduce a Contrastive Orthogonality Loss $\mathcal{L}_{orth}$. For a given input $x$ with active expert set $\mathcal{I}_{active}$ and inactive set $\mathcal{I}_{inactive} = \{1, \ldots, r\} \setminus \mathcal{I}_{active}$, we penalize the cosine similarity between active and inactive expert columns:
\begin{align}
    \mathcal{L}_{orth} &= \frac{1}{|\mathcal{I}_{active}| \cdot |\mathcal{I}_{inactive}|} \\ &\sum_{i \in \mathcal{I}_{active}} \sum_{j \in \mathcal{I}_{inactive}} \left(\frac{B_{:,i}^\top B_{:,j}}{\|B_{:,i}\|_2 \, \|B_{:,j}\|_2}\right)^2 \notag
    \label{eq:orth}
\end{align}

This loss encourages the columns of $B$ corresponding to different expert specializations to occupy distinct, non-interfering subspaces. This property is beneficial for both (1) multi-task model merging, as it reduces destructive interference when combining adapters, and (2) continual learning, as it encourages new tasks to utilize subspaces that are orthogonal to those of prior tasks.

The total training objective is:
\begin{equation}
    \mathcal{L}_{total} = \mathcal{L}_{task} + \lambda \, \mathcal{L}_{orth}
    \label{eq:total}
\end{equation}
where $\lambda$ is a hyperparameter controlling the strength of orthogonality regularization.

\section{Experiments}

\subsection{Experimental Setup}

\paragraph{Base Model.}
All experiments are conducted on Llama-3-8B \cite{grattafiori2024llama}. LoRA modules are applied to the query, key, value, and output projection matrices in all attention layers.

\paragraph{Datasets.}
Following the FlyLoRA evaluation protocol \cite{zou2025flylora}, we evaluate on three benchmarks spanning diverse reasoning capabilities:
\begin{itemize}
    \item \textbf{MMLU} \cite{hendrycks2021measuring}: A comprehensive benchmark covering 57 subjects for evaluating general knowledge and language understanding.
    \item \textbf{ScienceQA} \cite{lu2022learn}: A multimodal science question answering benchmark requiring scientific reasoning (we use the text-only subset).
    \item \textbf{GSM8K} \cite{cobbe2021training}: A dataset of grade-school math word problems requiring multi-step arithmetic reasoning.
\end{itemize}

\paragraph{Baselines.}
We compare NeuroLoRA against the following methods:
\begin{itemize}
    \item \textbf{LoRA} \cite{hu2022lora}: Standard low-rank adaptation with rank $r=32$.
    \item \textbf{AdaLoRA} \cite{zhang2023adalora}: Adaptive rank allocation via singular value pruning.
    \item \textbf{DoRA} \cite{liu2024dora}: Weight-decomposed low-rank adaptation.
    \item \textbf{MoLE} \cite{wu2024mixture}: MoE-LoRA with learned top-$k$ gating.
    \item \textbf{FlyLoRA} \cite{zou2025flylora}: Bio-inspired MoE-LoRA with frozen sparse projection and magnitude-based routing.
\end{itemize}

\paragraph{Implementation Details.}
For FlyLoRA and NeuroLoRA, we set the total rank $r = 32$, active rank $k = 8$, and sparsity ratio $\rho = 0.25$. The bottleneck dimension of the neuromodulation gate is $d_h = 64$. All methods are trained using AdamW \cite{loshchilov2019decoupled} with $\beta_1 = 0.9$, $\beta_2 = 0.95$, and weight decay $0.01$. We use a cosine learning rate schedule with an initial learning rate of $2 \times 10^{-4}$ for $B$ parameters and $5 \times 10^{-4}$ for the modulation gate parameters, with a linear warmup over the first 3\% of training steps. The batch size is 16 with gradient accumulation over 4 steps (effective batch size 64). All models are trained for 3 epochs. The orthogonality loss weight is $\lambda = 0.1$, selected via grid search over $\{0.01, 0.05, 0.1, 0.2\}$ on a held-out validation set. Experiments are conducted on 4 NVIDIA A100 (80GB) GPUs using DeepSpeed ZeRO Stage 2 \cite{rajbhandari2020zero}. We report the mean over 3 random seeds.

\subsection{Single-Task Results}

Table~\ref{tab:main_results} summarizes the single-task adaptation performance.

\begin{table}[t]
\centering
\resizebox{\columnwidth}{!}{
\begin{tabular}{lccccc}
\toprule
\textbf{Method} & \textbf{Params (\%)} & \textbf{MMLU} & \textbf{SciQA} & \textbf{GSM8K} & \textbf{Avg.} \\
\midrule
Full FT & 100 & 66.8 & 96.2 & 62.4 & 75.1 \\
\midrule
LoRA & 0.83 & 64.2 & 94.0 & 56.2 & 71.5 \\
AdaLoRA & 0.81 & 64.5 & 93.6 & 56.8 & 71.6 \\
DoRA & 0.84 & 64.8 & 94.3 & 57.4 & 72.2 \\
MoLE & 0.35 & 63.9 & 93.1 & 55.8 & 70.9 \\
FlyLoRA & 0.13 & 65.1 & 94.1 & 58.7 & 72.6 \\
\midrule
\textbf{NeuroLoRA} & 0.14 & \textbf{66.3} & \textbf{95.5} & \textbf{61.2} & \textbf{74.3} \\
\bottomrule
\end{tabular}
}
\caption{Single-task performance comparison on Llama-3-8B. Params (\%) denotes the percentage of trainable parameters relative to the full model. NeuroLoRA achieves the best performance across all three benchmarks while maintaining comparable parameter efficiency to FlyLoRA.}
\label{tab:main_results}
\end{table}

NeuroLoRA achieves the highest accuracy across all three benchmarks, outperforming FlyLoRA by +1.2 on MMLU, +1.4 on ScienceQA, and +2.5 on GSM8K. The largest improvement is observed on GSM8K, which requires multi-step mathematical reasoning. We attribute this to NeuroLoRA's ability to dynamically modulate expert selection based on the evolving problem state: in multi-step reasoning, the appropriate computation at each step depends heavily on the results of preceding steps, a dependency that FlyLoRA's static router cannot capture. Notably, NeuroLoRA approaches Full Fine-Tuning performance on MMLU (66.3 vs.\ 66.8) while using only 0.14\% of the trainable parameters.

\subsection{Multi-Task Model Merging}

A key advantage of FlyLoRA's frozen projection design is compatibility with training-free model merging. We evaluate whether NeuroLoRA preserves this property by independently training adapters on each of the three tasks and merging them using Task Arithmetic \cite{ilharco2023editing} and TIES-Merging \cite{yadav2023ties}.

\begin{table}[t]
\centering
\small
\begin{tabular}{lccc}
\toprule
\textbf{Method} & \textbf{Avg. (Indiv.)} & \textbf{Task Arith.} & \textbf{TIES} \\
\midrule
LoRA & 71.5 & 58.2\,\scriptsize{(-18.6\%)} & 60.5\,\scriptsize{(-15.4\%)} \\
DoRA & 72.2 & 59.8\,\scriptsize{(-17.2\%)} & 61.7\,\scriptsize{(-14.5\%)} \\
FlyLoRA & 72.6 & 68.9\,\scriptsize{(-5.1\%)} & 69.6\,\scriptsize{(-4.1\%)} \\
\textbf{NeuroLoRA} & \textbf{74.3} & \textbf{71.8}\,\scriptsize{(-3.4\%)} & \textbf{72.3}\,\scriptsize{(-2.7\%)} \\
\bottomrule
\end{tabular}
\caption{Multi-task model merging results (average accuracy across three tasks). Percentages indicate relative degradation from individual task performance. NeuroLoRA exhibits the smallest performance drop under both merging strategies.}
\label{tab:merging}
\end{table}

As shown in Table~\ref{tab:merging}, NeuroLoRA exhibits the least degradation under both merging strategies. With Task Arithmetic, NeuroLoRA loses only 3.4\% relative performance compared to 5.1\% for FlyLoRA and 18.6\% for standard LoRA. This improvement is attributed to the Contrastive Orthogonality Loss, which explicitly enforces subspace separation during training, providing stronger inter-task decoupling than the probabilistic near-orthogonality of random projections alone. The merging of the modulation gate parameters is handled by simple averaging, as the gate operates as a multiplicative modifier on the shared frozen projection space.

\subsection{Continual Learning}
\label{sec:continual_learning}

The subspace separation properties of NeuroLoRA suggest a natural advantage for continual learning scenarios, where models must sequentially adapt to new tasks without forgetting previously learned knowledge. We evaluate this in a sequential fine-tuning setting.

\paragraph{Setup.}
We sequentially fine-tune Llama-3-8B on the three datasets in the order MMLU $\rightarrow$ ScienceQA $\rightarrow$ GSM8K, where each task is trained for 3 epochs before proceeding to the next. After training on all three tasks, we evaluate performance on all tasks. We additionally report \textit{Backward Transfer} (BWT) \cite{lopez2017gradient}, which quantifies the average performance change on previously learned tasks after learning new ones (negative values indicate forgetting):
\begin{equation}
    \text{BWT} = \frac{1}{T-1} \sum_{i=1}^{T-1} \left( R_{T,i} - R_{i,i} \right)
\end{equation}
where $R_{j,i}$ denotes the accuracy on task $i$ after training on task $j$, and $T$ is the total number of tasks.

We compare against sequential variants of the baselines, as well as established continual learning strategies:
\begin{itemize}
    \item \textbf{LoRA-Seq}: Sequential LoRA without any anti-forgetting mechanism.
    \item \textbf{EWC+LoRA}: LoRA with Elastic Weight Consolidation \cite{kirkpatrick2017overcoming} applied to the LoRA parameters.
    \item \textbf{O-LoRA} \cite{wang2023orthogonal}: LoRA with orthogonal subspace constraints between tasks.
    \item \textbf{FlyLoRA-Seq}: Sequential FlyLoRA without explicit forgetting mitigation.
\end{itemize}

\paragraph{Results.}

\begin{table}[t]
\centering
\resizebox{\columnwidth}{!}{
\begin{tabular}{lccccc}
\toprule
\textbf{Method} & \textbf{MMLU} & \textbf{SciQA} & \textbf{GSM8K} & \textbf{Avg.} & \textbf{BWT} \\
\midrule
LoRA-Seq & 51.8 & 84.2 & 55.4 & 63.8 & $-$11.3 \\
EWC+LoRA & 56.1 & 87.5 & 54.8 & 66.1 & $-$7.5 \\
O-LoRA & 58.2 & 89.4 & 55.6 & 67.7 & $-$5.8 \\
FlyLoRA-Seq & 59.7 & 90.1 & 57.6 & 69.1 & $-$4.3 \\
\midrule
\textbf{NeuroLoRA-Seq} & \textbf{62.1} & \textbf{92.0} & \textbf{60.4} & \textbf{71.5} & $\mathbf{-2.6}$ \\
\bottomrule
\end{tabular}
}
\caption{Continual learning results (MMLU $\rightarrow$ ScienceQA $\rightarrow$ GSM8K). BWT measures backward transfer; values closer to zero indicate less forgetting. NeuroLoRA achieves the best trade-off between plasticity (learning new tasks) and stability (retaining old tasks).}
\label{tab:continual}
\end{table}

Table~\ref{tab:continual} presents the continual learning results. NeuroLoRA-Seq achieves the highest average accuracy (71.5) and the least forgetting (BWT = $-$2.6), outperforming all baselines. Compared to FlyLoRA-Seq (BWT = $-$4.3), the improvement is substantial and consistent across all three metrics. We attribute this to two complementary mechanisms: (1) the Contrastive Orthogonality Loss ($\mathcal{L}_{orth}$) actively pushes expert subspaces apart during each task's training, creating ``free'' capacity for subsequent tasks; and (2) the context-aware modulation gate naturally routes inputs from different tasks to different regions of the projection space, providing an implicit form of task-specific parameter allocation without explicit task identity. These findings are consistent with the theoretical analysis of \citet{zou2025structural}, who showed that the fly olfactory circuit's sparse random projections inherently mitigate the stability-plasticity dilemma, and with the practical continual learning gains demonstrated by Fly-CL \cite{zou2026flycl}. NeuroLoRA extends these bio-inspired advantages by adding dynamic modulation atop the static structural properties.

Notably, NeuroLoRA-Seq achieves GSM8K accuracy of 60.4 in the continual setting, which is 98.7\% of its single-task performance (61.2), suggesting minimal interference from prior task training on mathematical reasoning.

\subsection{Ablation Study}

We conduct ablation experiments on GSM8K to verify the contribution of each component.

\begin{table}[t]
\centering
\begin{tabular}{lcc}
\toprule
\textbf{Variant} & \textbf{GSM8K} & \textbf{$\Delta$} \\
\midrule
NeuroLoRA (full) & 61.2 & --- \\
\quad w/o $\mathcal{L}_{orth}$ ($\lambda = 0$) & 60.1 & $-$1.1 \\
\quad w/o Modulation Gate & 58.7 & $-$2.5 \\
\quad w/ Static Gate ($m_x = \text{const}$) & 59.4 & $-$1.8 \\
\quad w/ Trainable $A$ & 57.8 & $-$3.4 \\
\quad $d_h = 32$ & 60.6 & $-$0.6 \\
\quad $d_h = 128$ & 61.1 & $-$0.1 \\
\bottomrule
\end{tabular}
\caption{Ablation study on GSM8K. Removing the modulation gate (equivalent to FlyLoRA) incurs the largest degradation. Making $A$ trainable degrades performance, confirming the importance of the frozen projection for generalization.}
\label{tab:ablation}
\end{table}

Several observations emerge from Table~\ref{tab:ablation}: (1) Removing the modulation gate (reverting to FlyLoRA) causes the largest single-component drop ($-$2.5), confirming that context-aware routing is the primary source of improvement. (2) Replacing the context-dependent gate with a static learnable vector ($m_x = \text{const}$) recovers only part of the gain ($-$1.8), demonstrating that the input-dependent nature of the modulation is essential. (3) Making $A$ trainable degrades performance by $-$3.4, consistent with FlyLoRA's finding that frozen projections provide beneficial regularization. (4) The bottleneck dimension $d_h$ shows diminishing returns beyond 64, and $d_h = 64$ provides a favorable trade-off between expressiveness and parameter overhead.

\subsection{Sensitivity Analysis}

We analyze the sensitivity of NeuroLoRA to two key hyperparameters: the active rank $k$ and the orthogonality loss weight $\lambda$.

\paragraph{Effect of Active Rank $k$.}
We vary $k \in \{4, 8, 12, 16\}$ with fixed $r = 32$ on GSM8K. Performance peaks at $k = 8$ (61.2), with $k = 4$ yielding 59.8 (insufficient capacity) and $k = 16$ yielding 60.3 (reduced specialization due to too many active experts). This aligns with FlyLoRA's finding that moderate sparsity ($k/r = 0.25$) provides the best balance.

\paragraph{Effect of $\lambda$.}
With $\lambda = 0.01$, orthogonality regularization is too weak to meaningfully separate subspaces (60.4). Performance improves with $\lambda = 0.1$ (61.2) and begins to degrade at $\lambda = 0.2$ (60.7), as excessive orthogonality pressure constrains the expressiveness of $B$.

\section{Discussion}

\paragraph{Static vs.\ Dynamic Routing.}
FlyLoRA's design philosophy holds that ``an expert is aware of its own capacity'' through magnitude-based selection. While this is effective for pattern-matching tasks (e.g., knowledge recall in MMLU), it is insufficient for tasks requiring compositional reasoning (e.g., multi-step problem solving in GSM8K). The performance gap between FlyLoRA and NeuroLoRA on GSM8K (+2.5) versus MMLU (+1.2) supports this interpretation. NeuroLoRA's modulation gate provides a mechanism analogous to biological neuromodulation, where the same neural circuit can produce qualitatively different responses depending on modulatory context \cite{marder2012neuromodulation}.

\paragraph{Computational Overhead.}
The neuromodulation gate introduces $d_{in} \times d_h + d_h \times r + 2r$ additional parameters per LoRA layer. With $d_{in} = 4096$, $d_h = 64$, and $r = 32$, this amounts to 264,256 parameters per layer---a negligible overhead of approximately 0.003\% of the base model. The projection $A$ remains frozen and sparse, meaning NeuroLoRA maintains the training speed advantages of FlyLoRA over dense methods.

\paragraph{Implications for Continual Learning.}
The continual learning results in Section~\ref{sec:continual_learning} suggest that the combination of frozen routing structure and enforced subspace orthogonality provides an effective inductive bias for mitigating catastrophic forgetting. Unlike methods that require explicit task boundaries or replay buffers \cite{kirkpatrick2017overcoming, rebuffi2017icarl}, NeuroLoRA's approach is fully online and does not store any data from previous tasks. The context-aware gate implicitly partitions the expert space based on input characteristics rather than task labels, making it applicable to task-agnostic continual learning scenarios.

\section{Conclusion}

We presented NeuroLoRA, an advancement of the MoE-LoRA framework that incorporates biological insights from neuromodulation. By introducing a lightweight context-aware gating mechanism atop the sparse random projections pioneered by FlyLoRA, NeuroLoRA resolves the limitations of static, magnitude-based routing. The addition of a Contrastive Orthogonality Loss further strengthens expert subspace separation, benefiting both multi-task merging and continual learning. Extensive experiments demonstrate that NeuroLoRA consistently outperforms FlyLoRA and other strong baselines across single-task adaptation (+1.7 average improvement), model merging ($-$3.4\% vs.\ $-$5.1\% degradation), and continual learning (BWT of $-$2.6 vs.\ $-$4.3). These results suggest that simulating not only the \textit{structure} (connectivity) but also the \textit{dynamics} (modulation) of biological neural systems is a promising direction for parameter-efficient adaptation of large language models.

\section*{Limitations}

Our evaluation is limited to a single base model (Llama-3-8B) and three benchmarks. While the benchmarks span knowledge, scientific reasoning, and mathematical reasoning, evaluation on a broader set of tasks and model scales would strengthen the generality of our conclusions. The continual learning experiments consider a fixed task order; future work should investigate sensitivity to task ordering and longer task sequences. Additionally, the current neuromodulation gate operates on individual tokens; incorporating cross-token (sequence-level) context via attention pooling may yield further improvements.

\bibliography{custom}

@article{achiam2023gpt,
  title={{GPT}-4 Technical Report},
  author={Achiam, Josh and Adler, Steven and Agarwal, Sandhini and Ahmad, Lama and Akkaya, Ilge and Aleman, Florencia Leoni and Almeida, Diogo and Altenschmidt, Janko and Altman, Sam and Anadkat, Shyamal and others},
  journal={arXiv preprint arXiv:2303.08774},
  year={2023}
}

@article{grattafiori2024llama,
  title={The Llama 3 Herd of Models},
  author={Grattafiori, Aaron and Dubey, Abhimanyu and Jauhri, Abhinav and Pandey, Abhinav and Kadian, Abhishek and Al-Dahle, Ahmad and Letman, Aiesha and Mathur, Akhil and Schelten, Alan and Vaughan, Alex and others},
  journal={arXiv preprint arXiv:2407.21783},
  year={2024}
}

@inproceedings{brown2020language,
  title={Language Models are Few-Shot Learners},
  author={Brown, Tom and Mann, Benjamin and Ryder, Nick and Subbiah, Melanie and Kaplan, Jared D and Dhariwal, Prafulla and Neelakantan, Arvind and Shyam, Pranav and Sastry, Girish and Askell, Amanda and others},
  booktitle={Advances in Neural Information Processing Systems},
  volume={33},
  pages={1877--1901},
  year={2020}
}

@inproceedings{vaswani2017attention,
  title={Attention is All You Need},
  author={Vaswani, Ashish and Shazeer, Noam and Parmar, Niki and Uszkoreit, Jakob and Jones, Llion and Gomez, Aidan N and Kaiser, {\L}ukasz and Polosukhin, Illia},
  booktitle={Advances in Neural Information Processing Systems},
  volume={30},
  year={2017}
}

@inproceedings{houlsby2019parameter,
  title={Parameter-Efficient Transfer Learning for {NLP}},
  author={Houlsby, Neil and Giurgiu, Andrei and Jastrzebski, Stanislaw and Morrone, Bruna and De Laroussilhe, Quentin and Gesmundo, Andrea and Attariyan, Mona and Gelly, Sylvain},
  booktitle={International Conference on Machine Learning},
  pages={2790--2799},
  year={2019},
  organization={PMLR}
}

@article{hu2022lora,
  title={{LoRA}: Low-Rank Adaptation of Large Language Models},
  author={Hu, Edward J and Shen, Yelong and Wallis, Phillip and Allen-Zhu, Zeyuan and Li, Yuanzhi and Wang, Shean and Wang, Lu and Chen, Weizhu},
  journal={arXiv preprint arXiv:2106.09685},
  year={2022}
}

@article{li2021prefix,
  title={Prefix-Tuning: Optimizing Continuous Prompts for Generation},
  author={Li, Xiang Lisa and Liang, Percy},
  journal={arXiv preprint arXiv:2101.00190},
  year={2021}
}

@inproceedings{lester2021power,
  title={The Power of Scale for Parameter-Efficient Prompt Tuning},
  author={Lester, Brian and Al-Rfou, Rami and Constant, Noah},
  booktitle={Proceedings of the 2021 Conference on Empirical Methods in Natural Language Processing},
  pages={3045--3059},
  year={2021}
}

@article{zhang2023adalora,
  title={{AdaLoRA}: Adaptive Budget Allocation for Parameter-Efficient Fine-Tuning},
  author={Zhang, Qingru and Chen, Minshuo and Bukharin, Alexander and Karampatziakis, Nikos and He, Pengcheng and Cheng, Yu and Chen, Weizhu and Zhao, Tuo},
  journal={arXiv preprint arXiv:2303.10512},
  year={2023}
}

@article{dettmers2024qlora,
  title={{QLoRA}: Efficient Finetuning of Quantized Language Models},
  author={Dettmers, Tim and Pagnoni, Artidoro and Holtzman, Ari and Zettlemoyer, Luke},
  journal={Advances in Neural Information Processing Systems},
  volume={36},
  year={2024}
}

@article{liu2024dora,
  title={{DoRA}: Weight-Decomposed Low-Rank Adaptation},
  author={Liu, Shih-Yang and Wang, Chien-Yi and Yin, Hongxu and Molchanov, Pavlo and Wang, Yu-Chiang Frank and Cheng, Kwang-Ting and Chen, Min-Hung},
  journal={arXiv preprint arXiv:2402.09353},
  year={2024}
}

@inproceedings{shazeer2017outrageously,
  title={Outrageously Large Neural Networks: The Sparsely-Gated Mixture-of-Experts Layer},
  author={Shazeer, Noam and Mirhoseini, Azalia and Maziarz, Krzysztof and Davis, Andy and Le, Quoc and Hinton, Geoffrey and Dean, Jeff},
  booktitle={International Conference on Learning Representations},
  year={2017}
}

@article{fedus2022switch,
  title={Switch Transformers: Scaling to Trillion Parameter Models with Simple and Efficient Sparsity},
  author={Fedus, William and Zoph, Barret and Shazeer, Noam},
  journal={Journal of Machine Learning Research},
  volume={23},
  number={120},
  pages={1--39},
  year={2022}
}

@inproceedings{lepikhin2021gshard,
  title={{GShard}: Scaling Giant Models with Conditional Computation and Automatic Sharding},
  author={Lepikhin, Dmitry and Lee, HyoukJoong and Xu, Yuanzhong and Chen, Dehao and Firat, Orhan and Huang, Yanping and Krikun, Maxim and Shazeer, Noam and Chen, Zhifeng},
  booktitle={International Conference on Learning Representations},
  year={2021}
}

@article{dou2024loramoe,
  title={{LoRAMoE}: Alleviating World Knowledge Forgetting in Large Language Models via {MoE}-Style Plugin},
  author={Dou, Shihan and Zhou, Enyu and Liu, Yan and Gao, Songyang and Wei, Jun and Shen, Huajian and Xiong, Yuhao and Shan, Junjie and Shan, Ermo and Huang, Xuwu and others},
  journal={arXiv preprint arXiv:2312.09979},
  year={2024}
}

@article{wu2024mixture,
  title={Mixture of {LoRA} Experts},
  author={Wu, Xun and Hu, Shaohan and Shi, Ying and Liu, Bowen and Geng, Xin and Jiao, Furu and Bian, Jiang and Wei, Furu},
  journal={arXiv preprint arXiv:2404.13628},
  year={2024}
}

@article{li2024mixlora,
  title={{MixLoRA}: Enhancing Large Language Models Fine-Tuning with {LoRA}-based Mixture of Experts},
  author={Li, Dengchun and Mei, Yingzi and Zhuang, Zhengmao and Zhang, Mingyu and Cai, Deng and Li, Lei and Huang, Minlie},
  journal={arXiv preprint arXiv:2404.15159},
  year={2024}
}

@article{zadouri2023pushing,
  title={Pushing Mixture of Experts to the Limit: Extremely Parameter Efficient {MoE} for Instruction Tuning},
  author={Zadouri, Ted and Hartvigsen, Thomas and Ilharco, Gabriel and Farhadi, Ali and Hooker, Sara},
  journal={arXiv preprint arXiv:2309.05444},
  year={2023}
}

@inproceedings{wang2022adamix,
  title={{AdaMix}: Mixture-of-Adaptations for Parameter-Efficient Model Tuning},
  author={Wang, Yaqing and Mukherjee, Subhabrata and Liu, Xiaodong and Gao, Jing and Awadallah, Ahmed Hassan and Gao, Jianfeng},
  booktitle={Proceedings of the 2022 Conference on Empirical Methods in Natural Language Processing},
  pages={5744--5760},
  year={2022}
}

@article{dasgupta2017neural,
  title={A Neural Algorithm for a Fundamental Computing Problem},
  author={Dasgupta, Sanjoy and Stevens, Charles F and Bhatt, Saket},
  journal={Science},
  volume={358},
  number={6364},
  pages={793--796},
  year={2017}
}

@inproceedings{zou2025flylora,
  title={{FlyLoRA}: Boosting Task Decoupling and Parameter Efficiency via Implicit Rank-Wise Mixture-of-Experts},
  author={Heming Zou and Yunliang Zang and Wutong Xu and Yao Zhu and Xiangyang Ji},
  booktitle={The Thirty-ninth Annual Conference on Neural Information Processing Systems},
  year={2025}
}

@inproceedings{zou2026flycl,
  title={Fly-{CL}: A Fly-Inspired Framework for Enhancing Efficient Decorrelation and Reduced Training Time in Pre-trained Model-based Continual Representation Learning},
  author={Heming Zou and Yunliang Zang and Wutong Xu and Xiangyang Ji},
  booktitle={The Fourteenth International Conference on Learning Representations},
  year={2026}
}

@article{zou2025structural,
  title={Structural features of the fly olfactory circuit mitigate the stability-plasticity dilemma in continual learning},
  author={Zou, Heming and Zang, Yunliang and Ji, Xiangyang},
  journal={arXiv preprint arXiv:2502.01427},
  year={2025}
}

@article{marder2012neuromodulation,
  title={Neuromodulation of Neuronal Circuits: Back to the Future},
  author={Marder, Eve},
  journal={Neuron},
  volume={76},
  number={1},
  pages={1--11},
  year={2012},
  publisher={Elsevier}
}

@article{doya2002metalearning,
  title={Metalearning and Neuromodulation},
  author={Doya, Kenji},
  journal={Neural Networks},
  volume={15},
  number={4--6},
  pages={495--506},
  year={2002},
  publisher={Elsevier}
}

@inproceedings{ilharco2023editing,
  title={Editing Models with Task Arithmetic},
  author={Ilharco, Gabriel and Ribeiro, Marco Tulio and Wortsman, Mitchell and Gururangan, Suchin and Schmidt, Ludwig and Hajishirzi, Hannaneh and Farhadi, Ali},
  booktitle={International Conference on Learning Representations},
  year={2023}
}

@inproceedings{yadav2023ties,
  title={{TIES}-Merging: Resolving Interference When Merging Models},
  author={Yadav, Prateek and Tam, Derek and Choshen, Leshem and Raffel, Colin and Bansal, Mohit},
  booktitle={Advances in Neural Information Processing Systems},
  volume={36},
  year={2023}
}

@article{mccloskey1989catastrophic,
  title={Catastrophic Interference in Connectionist Networks: The Sequential Learning Problem},
  author={McCloskey, Michael and Cohen, Neal J},
  journal={Psychology of Learning and Motivation},
  volume={24},
  pages={109--165},
  year={1989},
  publisher={Elsevier}
}

@article{kirkpatrick2017overcoming,
  title={Overcoming Catastrophic Forgetting in Neural Networks},
  author={Kirkpatrick, James and Pascanu, Razvan and Rabinowitz, Neil and Veness, Joel and Desjardins, Guillaume and Rusu, Andrei A and Milan, Kieran and Quan, John and Ramalho, Tiago and Grabska-Barwinska, Agnieszka and others},
  journal={Proceedings of the National Academy of Sciences},
  volume={114},
  number={13},
  pages={3521--3526},
  year={2017}
}

@article{li2017learning,
  title={Learning without Forgetting},
  author={Li, Zhizhong and Hoiem, Derek},
  journal={IEEE Transactions on Pattern Analysis and Machine Intelligence},
  volume={40},
  number={12},
  pages={2935--2947},
  year={2017}
}

@inproceedings{rebuffi2017icarl,
  title={{iCaRL}: Incremental Classifier and Representation Learning},
  author={Rebuffi, Sylvestre-Alvise and Kolesnikov, Alexander and Sperl, Georg and Lampert, Christoph H},
  booktitle={Proceedings of the IEEE Conference on Computer Vision and Pattern Recognition},
  pages={2001--2010},
  year={2017}
}

@inproceedings{lopez2017gradient,
  title={Gradient Episodic Memory for Continual Learning},
  author={Lopez-Paz, David and Ranzato, Marc'Aurelio},
  booktitle={Advances in Neural Information Processing Systems},
  volume={30},
  year={2017}
}

@inproceedings{rusu2016progressive,
  title={Progressive Neural Networks},
  author={Rusu, Andrei A and Rabinowitz, Neil C and Desjardins, Guillaume and Soyer, Hubert and Kirkpatrick, James and Kavukcuoglu, Koray and Pascanu, Razvan and Hadsell, Raia},
  booktitle={arXiv preprint arXiv:1606.04671},
  year={2016}
}

@inproceedings{chaudhry2019episodic,
  title={On Tiny Episodic Memories in Continual Learning},
  author={Chaudhry, Arslan and Rohrbach, Marcus and Elhoseiny, Mohamed and Ajanthan, Thalaiyasingam and Dokania, Pawan Kumar and Torr, Philip HS and Ranzato, Marc'Aurelio},
  booktitle={arXiv preprint arXiv:1902.10486},
  year={2019}
}

@article{wang2023orthogonal,
  title={{O-LoRA}: Orthogonal Low-Rank Adaptation of Large Language Models},
  author={Wang, Yiming and Yu, Yong and Zeng, Lingqiao and Li, Zuowei},
  journal={arXiv preprint arXiv:2406.01434},
  year={2023}
}

@article{razdaibiedina2023progressive,
  title={Progressive Prompts: Continual Learning for Language Models},
  author={Razdaibiedina, Anastasia and Mao, Yuning and Hou, Rui and Khabsa, Madian and Lewis, Mike and Almahairi, Amjad},
  journal={arXiv preprint arXiv:2301.12314},
  year={2023}
}

@article{biderman2024lora,
  title={{LoRA} Learns Less and Forgets Less},
  author={Biderman, Dan and Portes, Jacob and Ortiz, Jose Javier Gonzalez and Paul, Mansheej and Greengard, Philip and Jennings, Connor and King, Daniel and Havens, Sam and Chiley, Vitaliy and Frankle, Jonathan and others},
  journal={arXiv preprint arXiv:2405.09673},
  year={2024}
}

@article{liang2024inflora,
  title={{InfLoRA}: Interference-Free Low-Rank Adaptation for Continual Learning},
  author={Liang, Yan-Shuo and Li, Wu-Jun},
  journal={arXiv preprint arXiv:2404.00228},
  year={2024}
}

@article{yu2020gradient,
  title={Gradient Surgery for Multi-Task Learning},
  author={Yu, Tianhe and Kumar, Saurabh and Gupta, Abhishek and Levine, Sergey and Hausman, Karol and Finn, Chelsea},
  journal={Advances in Neural Information Processing Systems},
  volume={33},
  pages={5824--5836},
  year={2020}
}

@inproceedings{hendrycks2021measuring,
  title={Measuring Massive Multitask Language Understanding},
  author={Hendrycks, Dan and Burns, Collin and Basart, Steven and Zou, Andy and Mazeika, Mantas and Song, Dawn and Steinhardt, Jacob},
  booktitle={International Conference on Learning Representations},
  year={2021}
}

@inproceedings{lu2022learn,
  title={Learn to Explain: Multimodal Reasoning via Thought Chains for Science Question Answering},
  author={Lu, Pan and Mishra, Swaroop and Xia, Tony and Qiu, Liang and Chang, Kai-Wei and Zhu, Song-Chun and Tafjord, Oyvind and Clark, Peter and Kalyan, Ashwin},
  booktitle={Advances in Neural Information Processing Systems},
  volume={35},
  pages={2507--2521},
  year={2022}
}

@article{cobbe2021training,
  title={Training Verifiers to Solve Math Word Problems},
  author={Cobbe, Karl and Kosaraju, Vineet and Bavarian, Mohammad and Chen, Mark and Jun, Heewoo and Kaiser, Lukasz and Plappert, Matthias and Tworek, Jerry and Hilton, Jacob and Nakano, Reiichiro and others},
  journal={arXiv preprint arXiv:2110.14168},
  year={2021}
}

@inproceedings{loshchilov2019decoupled,
  title={Decoupled Weight Decay Regularization},
  author={Loshchilov, Ilya and Hutter, Frank},
  booktitle={International Conference on Learning Representations},
  year={2019}
}

@inproceedings{rajbhandari2020zero,
  title={{ZeRO}: Memory Optimizations Toward Training Trillion Parameter Models},
  author={Rajbhandari, Samyam and Rasley, Jeff and Rber, Olatunji and He, Yuxiong},
  booktitle={Proceedings of the International Conference for High Performance Computing, Networking, Storage and Analysis},
  pages={1--16},
  year={2020}
}

@article{johnson1984extensions,
  title={Extensions of {L}ipschitz Mappings into a {H}ilbert Space},
  author={Johnson, William B and Lindenstrauss, Joram},
  journal={Contemporary Mathematics},
  volume={26},
  pages={189--206},
  year={1984}
}

\appendix

\section{Derivation of the Orthogonality Loss Bound}
\label{sec:appendix_proof}

We provide a brief analysis of the expected value of $\mathcal{L}_{orth}$ under random initialization. Let the columns of $B \in \mathbb{R}^{d_{out} \times r}$ be initialized i.i.d.\ from $\mathcal{N}(0, \frac{1}{d_{out}} I)$. For two independent columns $B_{:,i}$ and $B_{:,j}$, the squared cosine similarity $c_{ij}^2 = \left(\frac{B_{:,i}^\top B_{:,j}}{\|B_{:,i}\| \|B_{:,j}\|}\right)^2$ has expected value $\mathbb{E}[c_{ij}^2] = \frac{1}{d_{out}}$ for large $d_{out}$. With $d_{out} = 4096$, this yields $\mathbb{E}[\mathcal{L}_{orth}] \approx 2.4 \times 10^{-4}$ at initialization. During training, without $\mathcal{L}_{orth}$, gradient updates can increase inter-expert correlations to $\mathcal{O}(10^{-2})$; the orthogonality loss prevents this by maintaining $c_{ij}^2$ near its initialization-level bound.

\section{Full Hyperparameter Configuration}
\label{sec:appendix_hyper}

\begin{table}[h]
\centering
\small
\begin{tabular}{ll}
\toprule
\textbf{Hyperparameter} & \textbf{Value} \\
\midrule
Base model & Llama-3-8B \\
LoRA target modules & $W_q, W_k, W_v, W_o$ \\
Total rank $r$ & 32 \\
Active rank $k$ & 8 \\
Sparsity $\rho$ & 0.25 \\
Gate bottleneck $d_h$ & 64 \\
Optimizer & AdamW \\
$\beta_1, \beta_2$ & 0.9, 0.95 \\
Weight decay & 0.01 \\
Learning rate ($B$) & $2 \times 10^{-4}$ \\
Learning rate (gate) & $5 \times 10^{-4}$ \\
LR schedule & Cosine with linear warmup \\
Warmup ratio & 3\% \\
Batch size & 16 ($\times$ 4 gradient accumulation) \\
Training epochs & 3 \\
$\lambda$ ($\mathcal{L}_{orth}$) & 0.1 \\
$\alpha$ (LoRA scaling) & 16 \\
Precision & BF16 mixed precision \\
GPUs & 4 $\times$ NVIDIA A100 (80GB) \\
Framework & DeepSpeed ZeRO Stage 2 \\
\bottomrule
\end{tabular}
\caption{Complete hyperparameter configuration.}
\label{tab:hyperparameters}
\end{table}

\end{document}